\documentclass{INTERSPEECH2023}

\interspeechcameraready

\title{SpellMapper: A non-autoregressive  neural spellchecker for ASR customization with candidate retrieval based on n-gram mappings}
\name{Alexandra Antonova, Evelina Bakhturina, Boris Ginsburg\sthanks{*Corresponding author}}
\address{
  NVIDIA, USA}
\email{\{aleksandraa, ebakhturina, bginsburg\}@nvidia.com}

\begin{document}

\maketitle
 
\begin{abstract}
Contextual spelling correction models are an alternative to shallow fusion to improve automatic speech recognition (ASR) quality given user vocabulary. To deal with large user vocabularies, most of these models include candidate retrieval mechanisms, usually based on minimum edit distance between fragments of ASR hypothesis and user phrases. However, the edit-distance approach is slow, non-trainable, and may have low recall as it relies only on common letters. We propose: 1) a novel algorithm for candidate retrieval, based on misspelled n-gram mappings, which gives up to 90\% recall with just the top 10 candidates on Spoken Wikipedia; 2) a non-autoregressive neural model based on BERT architecture, where the initial transcript and ten candidates are combined into one input. The experiments on Spoken Wikipedia show 21.4\% word error rate improvement compared to a baseline ASR system.

\end{abstract}

\section{Introduction}

Customization (contextual biasing) of automatic speech recognition (ASR) systems with user-defined phrases, such as proper names or terms, is required by many production systems. 

Contextual\footnote{We use the term ``context" for a user-defined list of phrases} spelling correction(CSC) is a promising approach to ASR customization that relies on postprocessing of ASR output and does not require retraining of the acoustic model.  
Unlike traditional ASR spellchecking approaches such as \cite{Guo2019ASC, hrinchuk2020correction}, which aim to correct known words using language models, the goal of contextual spelling correction is to correct highly specific user terms, most of which can be 1) out-of-vocabulary (OOV) words, 2) spelling variations (e.g., ``John Koehn", ``Jon Cohen") and language models cannot help much with that. 

Our approach and task formulation are inspired by the non-autoregressive model proposed in \cite{Wang2022TowardsCS}. The contextual spellchecking module takes as input a single ASR hypothesis and a user-defined vocabulary with up to several thousand words or phrases. The goal is to identify if a given ASR hypothesis contains any phrases from the user vocabulary, \textit{possibly in a misspelled form}, and their exact positions in text. Next, the model corrects the ASR hypothesis by replacing the misspelled fragment with a matched phrase from the user vocabulary. For example, given an ASR-hypothesis \textit{``came from schleiddorf in wurteenberg"} and user vocabulary including \{\textit{``schlaitdorf"}, \textit{``wurttemberg"}\}, the contextual spellchecking module should produce \textit{``came from schlaitdorf in wurttemberg"}.

CSC models rely on attention mechanisms to compare ASR hypothesis and candidate phrases. Unfortunately, the attention fails on large user vocabularies, and reducing the number of candidates is necessary. 
For example, in \cite{Wang2022TowardsCS}, the authors retrieve 100 candidates via minimum edit distance. They had to introduce a few custom blocks to their model architecture to support such a large number of candidates. The candidate retrieval approach based on edit distance suffers from two serious disadvantages. First, it is computationally inefficient due to quadratic complexity. Second, it heavily relies on common letters, which is not always enough, e.g., ``lookez" vs. ``lucas" have only one letter in common. 

\begin{figure*}[t]
    \centering
    \includegraphics[width=0.9\textwidth]{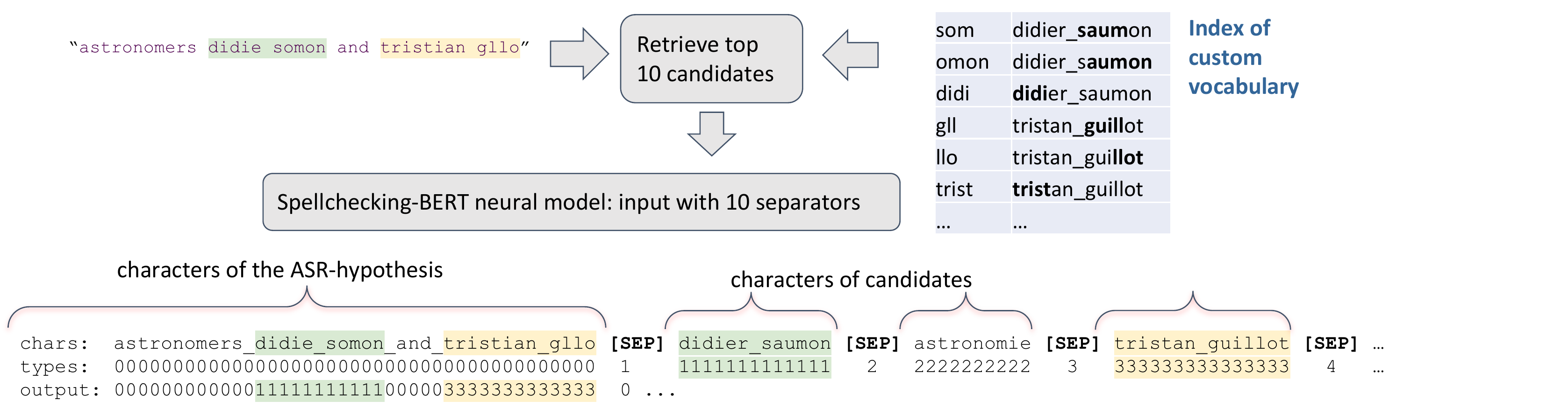}
    \caption {Inference pipeline. Candidate retrieval ingests ASR output fragment of 10-15 words and selects top 10 user phrases with many matching n-grams in the user vocabulary index. ASR-hypothesis and candidates are combined to a single BERT input sequence with separators. The model predicts labels of correct candidates for each character of ASR-hypothesis, or 0.}
\label{fig:inference}
\end{figure*}
We propose a novel algorithm for candidate retrieval based on misspelled n-gram mappings to overcome both disadvantages of the edit distance approach. It preselects the top 10 candidates from the user vocabulary, potentially matching the ASR hypothesis. All the retrieved candidates are then concatenated with the ASR output and passed to a non-autoregressive spellchecker based on BERT architecture\cite{DevlinCLT19}. The proposed approach requires two offline preparation steps: a collection of n-gram mappings from a large corpus and an indexing of custom vocabulary for a new or updated user. 

The contributions of the paper are the following:
\begin{enumerate}
\item A novel candidate retrieval algorithm based on misspelled n-gram mappings. It provides 74-90\% recall with just the top 10 candidates. 
\item A new way to formulate customization spellchecking as a non-autoregressive learning task for BERT architecture, using multiple separators (see Fig. \ref{fig:inference}).
\item A procedure to synthetically generate training datasets.
\end{enumerate}
The proposed spellchecking approach demonstrates sustainable improvement (6.4-21.4\% relative WER decrease) on three public datasets from different domains, for which we simulate ``user vocabularies".
The code is publicly released as part of NeMo toolkit \footnote{\url{https://github.com/NVIDIA/NeMo}}. 

\section{Background}
Existing approaches to text-only contextual biasing of end-to-end (E2E) ASR systems can be divided into four groups. \textit{TTS augmentation} is often used to fine-tune on-device ASR models with user text-only data \cite{sim2019personalization}. However, this approach is impractical for ASR done in the cloud because it requires retraining of the ASR model. \textit{Shallow fusion} \cite{Zhao2019ShallowFusionEC, Le2020DeepSF, gourav2021personalization, fox2022improving}  is based on rescoring paths corresponding to custom phrases in the recognition graph. This method is prone to false positives and is not applicable if the needed path is not present in the graph, which is often the case with end-to-end ASR.
\textit{Deep contextualization}  \cite{Pundak2018DeepCE,Jain2020ContextualRF} adds a custom vocabulary as an additional input to the acoustic model and tries to combine it with the audio input via a context encoder and some kind of attention. This approach preserves the E2E model nature and does not require model fine-tuning for each user. The primary constraint of this attention-based approach is that attention cannot handle large vocabularies, and candidate retrieval is not possible as ASR hypotheses are not known beforehand. These drawbacks can be resolved with a more complex neural architecture or switching to a two-stage approach with candidate retrieval for the second phase using ASR output from the first phase \cite{Yang2023TwoSC, Munkhdalai2023NAMTS}. \textit{Contextual spelling correction} requires a separate model that works on top of the ASR system \cite{wang2021light, Wang2022TowardsCS}. It detects and locates custom phrases that could be present in the ASR output text, possibly in a misrecognized form. This approach allows retrieval of a \textit{subset} of custom phrases based on the known ASR hypothesis. 

\begin{table}[b]
  \caption{N-gram mappings are pairs of n-grams and their common misspelled counterparts from the corpus automatically aligned on character level}
  \label{tab:ngram_mappings}
  \centering
  \begin{tabular}{rlr}
    \toprule
    \textbf{Source} & \textbf{Target} & \textbf{Frequency} \\
    \midrule
     l u c  &  l u c  &   7432 \\
     l u c  &  l u k+e  &  565  \\
     l u c  &  l o c  &   561  \\
     l u c  &  l u k  &   382  \\
     l u c  &  l o+o k  &   298  \\
     l u c  &  l u s  &  224  \\
     l u c  &  l u \textless{}DEL\textgreater{}  &  195  \\
    \bottomrule
  \end{tabular}
\end{table}

\section{Proposed approach}

\begin{figure}[t]
  \centering
  \includegraphics[width=0.9\linewidth]{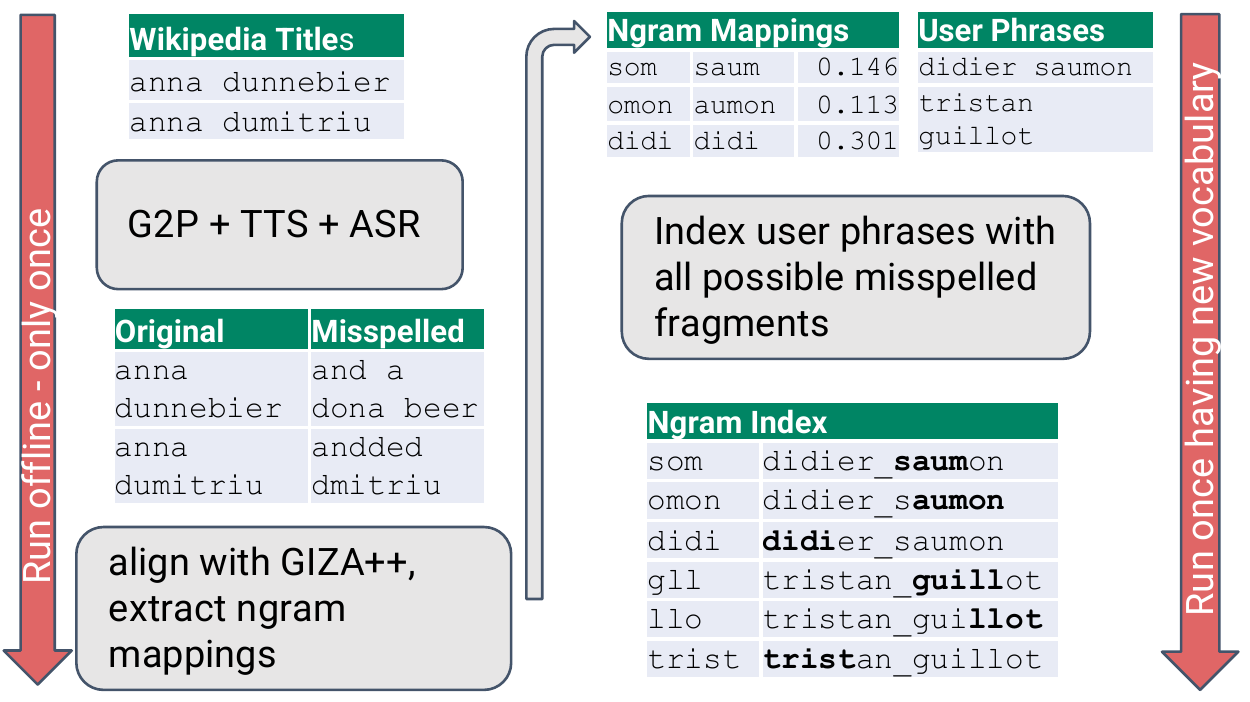}
  \caption{Data preparation pipeline. Left: N-gram mappings collection. Right: Indexing of user vocabulary.}
  \label{fig:data_preparation}
\end{figure}

Our approach to customization spellchecking consists of the following online modules: candidate retrieval, non-autoregressive neural model, and post-processing. The inference pipeline is shown in Fig. \ref{fig:inference}. It ingests ASR output fragments of 10-15 words and an n-gram-indexed user vocabulary. Then it retrieves the top 10 candidate phrases from the user vocabulary and tags the input characters with correct candidate labels or 0 if no match is found. The post-processing step combines outputs from different fragments, applies additional filtering, and outputs the final corrected ASR transcription. The proposed approach also requires two offline preparation steps: 1) collecting n-gram mappings from a large corpus and 2) vocabulary indexing for each user (Fig. \ref{fig:data_preparation}). 

\subsection{N-gram mappings and indexing of user vocabulary} \label{ngram_mappings_and_indexing}
\cite{brill-moore-2000-improved} proposed one of the first statistical models to learn probabilities of minor string-to-string edits from a corpus of paired misspelled and correct words. Following this, we collect n-gram mappings (see Table \ref{tab:ngram_mappings}) from a parallel corpus of Wikipedia headings and their corrupted version. To get the corrupted version, we synthesize audio from vocabulary phrases and then pass through an ASR model (Conformer-CTC, see Table \ref{tab:models}) to get text with potential errors. We align ground truth with corrupted phrases on a character level with GIZA++ \cite{och2003systematic}. Finally, we extract aligned n-grams from GIZA++ alignments along with their frequencies. We collect n-gram mappings from a large corpus only once.

Each time we have a new or updated user vocabulary we build an index where \textit{key} is a letter n-gram and \textit{value} is the whole phrase. The keys are n-grams in the given user phrase and their misspelled variants taken from our collection of n-gram mappings (see Index of custom vocabulary in Fig. \ref{fig:inference}).

\begin{figure}[t]
  \centering
  \includegraphics[width=\linewidth]{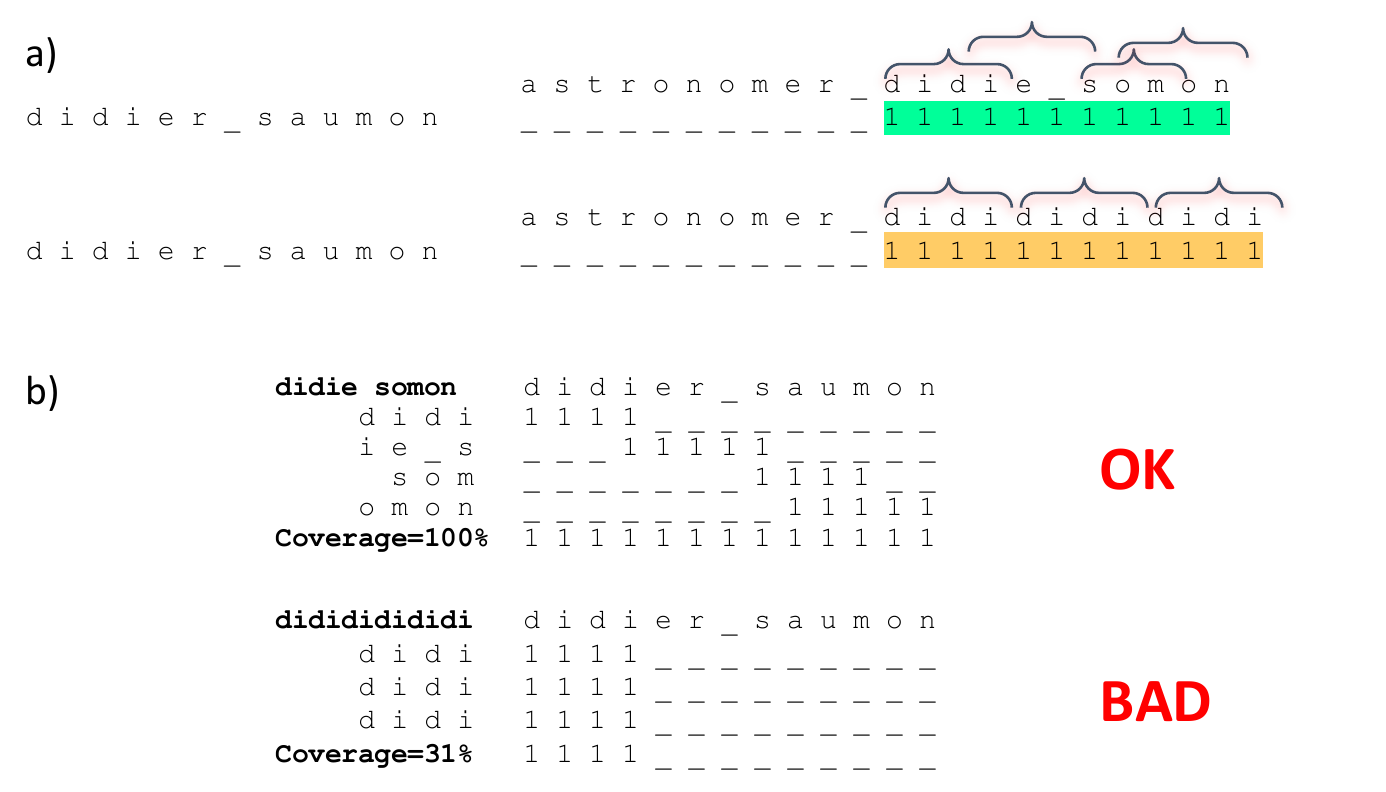}
  \caption{a) Detection of potential candidate phrases in the ASR hypothesis using n-gram index. Note that repetitive or similar n-grams can produce wrong candidates. b) Additional filtering of found candidates by checking how well the phrase is covered by matched ngrams.}
  \label{fig:detection_by_coverage}
\end{figure}

\subsection{Candidate retrieval algorithm}
The candidate retrieval algorithm works online before the neural model inference.
First, we search all overlapping n-grams in the index (each n-gram gives hits to corresponding symbols from ASR hypothesis) and select candidate phrases with the most hits. Next, we additionally filter out candidates by checking how well matched n-grams cover the phrase itself (Fig. \ref{fig:detection_by_coverage})

\subsection{Neural model} \label{neural_model}

The SpellMapper is based on BERT architecture \cite{DevlinCLT19}. The standard BERT input sequence consists of up to 512 tokens and can be segmented into subsequences of different types by inserting a special [SEP] token between input parts and by specifying additional input tensor \textit{token\_type\_ids}. Though this is not common to split the input into more than two segments(e.g., first and second sentences to compare, or short paragraph and question for question-answering problem) and none of pre-trained models were trained on such tasks, for SpellMapper we use ten separators to combine the ASR hypothesis and ten candidate phrases into a single input (see Fig. \ref{fig:inference})


The nature of spellchecking implies that we want to pay more attention to how fragments are similar letter-by-letter than the meaning of the words. To achieve this, we tokenize the input sequence into individual characters, replacing the space symbol with an underscore. The SpellMapper learns new character embeddings during training. 
We enhance our model with information from subword embeddings by passing the same input through standard BERT tokenization and concatenating subword embeddings to the embeddings of \textbf{each} character that belongs to this subword. 
While the model works on character level, adding subword embeddings may help the model better understand the context and reduce false positives by a quarter.

\subsection{Metrics} \label{metrics}
We use word error rate (WER) to compare transcripts corrected with SpellMapper and the baseline ASR output. 
We also develop a procedure that imitates the ``ideal spellchecker" to estimate maximum WER improvement with the given vocabulary, since the post-correction results highly depend on the vocabulary properties, e.g., size, minimum length, word frequencies, and their occasional similarity to frequent phrases. The ``ideal spellchecker" first aligns reference and baseline ASR transcription to get nonidentical fragments. If a reference fragment exists in the custom vocabulary, we count it as successfully corrected. The resulting ``ideal" WER gives us a lower bound on the best potential improvement with the given vocabulary.

Additionally, we calculate \textit{diff-keyword-dependent} precision and recall following \textit{keyword-dependent} metrics from \cite{sim2019personalization}. We do not count occurrences where the correct user phrase was present in both the original and corrected ASR hypothesis
(80-85\% in our experiments), and only consider cases where was some change, counting \textit{better}, \textit{missed} and \textit{false positive(fp)} cases. 
Recall (\ref{recall}) measures the model's ability to identify user phrases disregarding that ASR can misrecognize them, e.g., \textit{``time we of derby"} $=>$ \textit{``tyne wear derby"}. Precision (\ref{prec}) is affected by the ratio of false positives - hallucinating user phrases in the unrelated text, e.g., \textit{an artist's right to fail} $=>$ \textit{an artist's reilly to fail}.
\begin{equation}
\label{recall}
    Recall = \frac{better}{better+missed}
\end{equation}

\begin{equation}
\label{prec}
    Precision = \frac{better}{better+fp}
\end{equation}
We also calculate \textit{Top 10 recall} for the candidates produced by the retrieval algorithm. Its comparison with final recall reveals how many correct candidates were lost at the retrieval stage versus lost at the stage of neural model or post-processing.




\begin{table*}[th]
  \caption{Spellchecking results w.r.t. 3 datasets and 2 baselines Conformer models (Table \ref{tab:models}). Spellchecking gives 6.4-21.4\% relative WER improvement and allows to regain 49-68\% of maximum ``ideal" WER improvement. The metrics are described in Sec. \ref{metrics}. }
  \label{tab:spellchecking_results}
  \centering
  \begin{tabular}{rrrrrrrrrrr}
    \toprule
    \textbf{Dataset} & \textbf{Users} & \multicolumn{1}{c}{\begin{tabular}[c]{@{}c@{}}\textbf{Utte-}\\ \textbf{rances}\end{tabular}} & \multicolumn{1}{c}{\begin{tabular}[c]{@{}c@{}}\textbf{Max/Avg}\\ \textbf{vocab size}\end{tabular}} & \multicolumn{1}{c}{\begin{tabular}[c]{@{}c@{}}\textbf{ASR}\\ \textbf{Model}\end{tabular}} & \multicolumn{1}{c}{\begin{tabular}[c]{@{}c@{}}\textbf{Baseline}\\ \textbf{WER \%}\end{tabular}} & \multicolumn{1}{c}{\begin{tabular}[c]{@{}c@{}}\textbf{Spellcheck}\\ \textbf{WER \%}\end{tabular}} & \multicolumn{1}{c}{\begin{tabular}[c]{@{}c@{}}\textbf{``Ideal"}\\ \textbf{WER \%}\end{tabular}} & \multicolumn{1}{c}{\begin{tabular}[c]{@{}c@{}}\textbf{Recall}\\ \textbf{\%}\end{tabular}} & \multicolumn{1}{c}{\begin{tabular}[c]{@{}c@{}}\textbf{Precision}\\ \textbf{\%}\end{tabular}} & \multicolumn{1}{c}{\begin{tabular}[c]{@{}c@{}} \textbf{Top 10} \\ \textbf{recall \%} \end{tabular}} \\
  \midrule
     SWC  &  1341 & 61370  &   1341/172 & CTC & 6.69 & 5.26 & 4.46 & 67 & 87.4 & 90.4 \\
          &  & &   & RNNT & 6.07 & 4.93 & 4.12 & 62.9 & 85.6 & 88.1 \\
  \midrule
     SPGI  &  1114 & 39341  &   86/21  & CTC & 5.88 & 5.50 & 5.33 & 55 & 88.7 & 84.5 \\
          & &  &   & RNNT & 5.71 & 5.32 & 5.12 & 53.2 & 88.5 & 82.7 \\
  \midrule
     UserLibri  &  99 & 5559 &  192/43  & CTC & 3.35 & 2.89 & 2.52 & 61.7 & 82.3 & 83.9 \\
          &  & &  & RNNT & 2.77 & 2.44 & 2.09 & 54.5 & 77.6 & 74.1 \\
    \bottomrule
  \end{tabular}
\end{table*}

\begin{table}[t]
\caption{Pre-trained models used in this work.}
\label{tab:models}
\centering
\resizebox{\columnwidth}{!}{
\begin{tabular}{ll}
\toprule
\multicolumn{1}{l}{\textbf{Model name}}       & \textbf{Source}                                                \\ \midrule
\multicolumn{2}{l}{\textit{from \url{https://huggingface.co:}}}                               \\ 
\multicolumn{1}{l}{BERT}                      & huawei-noah/TinyBERT\_General\_6L\_768D \\ \midrule
\multicolumn{2}{l}{\textit{from  \url{https://catalog.ngc.nvidia.com/models:}}}                               \\ 
\multicolumn{1}{l}{Conformer-CTC}                   & stt\_en\_conformer\_ctc\_large                                 \\ 
\multicolumn{1}{l}{Conformer-RNNT}                  & stt\_en\_conformer\_transducer\_large                          \\ 
\multicolumn{1}{l}{Mel-spectrogram} & tts\_en\_fastpitch                     \\ 
\multicolumn{1}{l}{Vocoder}                   & tts\_hifigan                            \\ \bottomrule
\end{tabular}}
\label{tab:my-table}
\end{table}

\section{Experiments}
\subsection{Datasets} \label{datasets}
\subsubsection{Corpus for n-gram mappings} \label{corpus_of_ngram_mappings}
For the collection of n-gram mappings, we use a corpus of 4.5M Wikipedia headings from the Yago database \cite{pellissier2020yago}. We synthesize audio for each heading with TTS using FastPitch mel-spectrogram generator \cite{lancucki2021fastpitch} and HiFi-GAN vocoder\cite{kong2020hifi} (Table \ref{tab:models}). Then we transcribe all generated audios with an ASR model (Conformer-CTC), thus getting a ``parallel" corpus. The resulting corpus contains many examples of possible misspellings with WER of about 80\%. Next, we run GIZA++ alignment and extract 1.9M n-gram mappings with a maximum length of five characters and conditional probability $>$ 0.018.

\subsubsection{Training and evaluation datasets} 
\label{testing_datasets}
 Each training example includes an ASR hypothesis, ten candidate phrases, and target labels. We create this dataset synthetically, reusing our collected corpus of correct and misspelled Wikipedia titles (Section \ref{corpus_of_ngram_mappings}). We extract all sentence fragments from full Wikipedia articles, where any Wikipedia title occurred in a real context. Next, we substitute the correct spelling with a misspelled one (to imitate the ASR hypothesis). We use the original spelling as the correct candidate and sample incorrect candidates from random, similar, or intersecting titles. We also add 50\% of examples without any misspellings.

For the final testing, we prepare ``customization" datasets from three public corpora which provide a grouping of utterances per user or document: Spoken Wikipedia \cite{Baumann2019TheSW}, SPGISpeech validation part \cite{oneillSPGISpeech}, UserLibri \cite{BreinerRVGMSGCM22}.
For Spoken Wikipedia (SWC) we perform sentence splitting, text normalization, and segmentation as described in \cite{bakhturina2021toolbox}. 
For SPGISpeech we delete filler words, e.g., ``uh" and ``um", from baseline transcriptions as they are absent in reference. For each ``user", we select capitalized multi-word phrases and rare words from their reference texts to serve as custom vocabulary. UserLibri does not retain information about the original case, so only single custom words are extracted.


\subsection{Training and Testing procedures}
\begin{figure}[t]
  \centering
\includegraphics[height=0.65\linewidth]{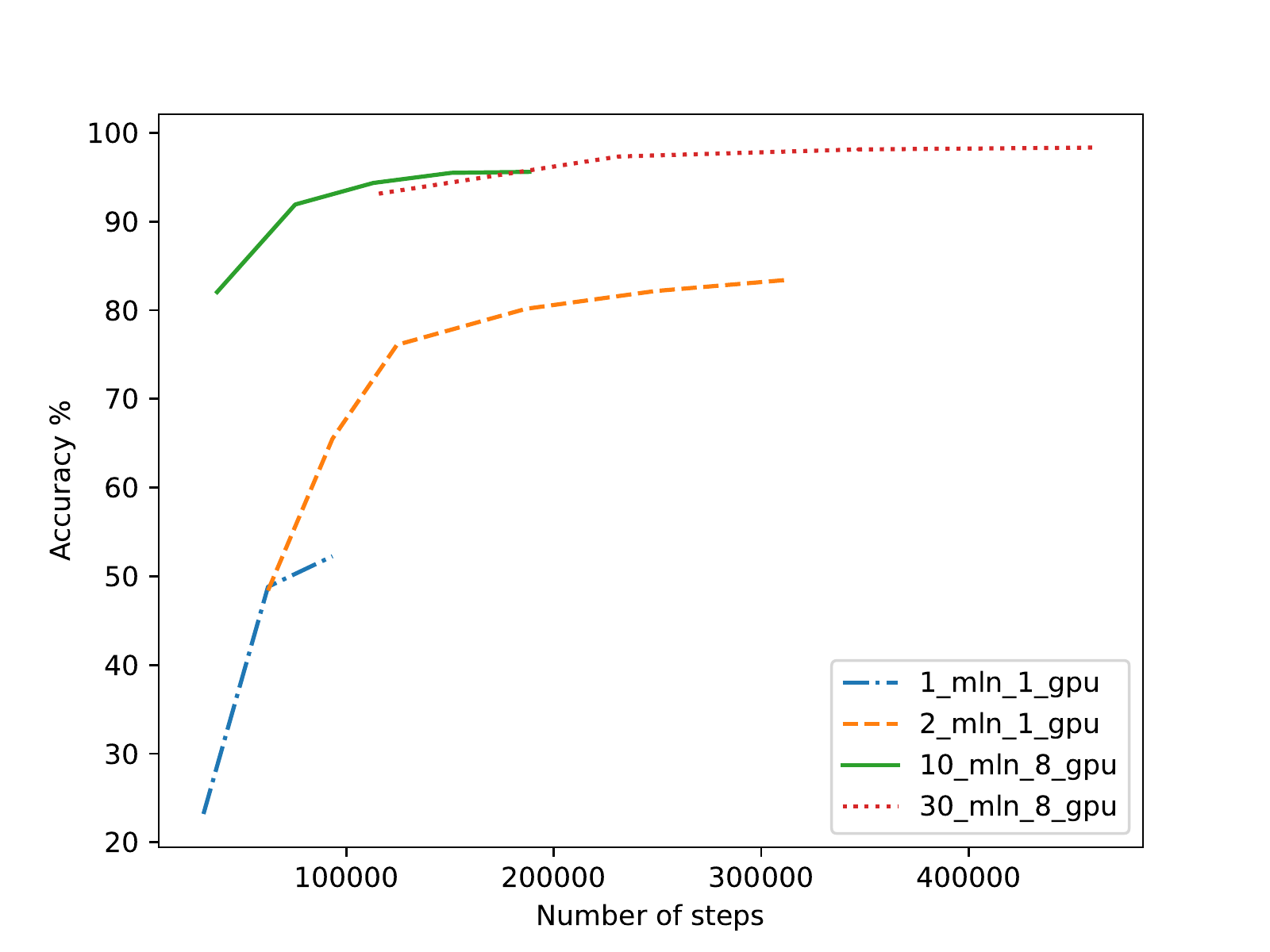}
  \caption{SpellMapper accuracy on classes 1-10 depending on the number of training examples (1, 2, 10, 30M) evaluated on 20K validation examples. The whole phrase is considered correct if all symbols are tagged correctly. On class 0 (not presented on the graph) accuracy per symbol is 93\%, 95.9\%, 97.3\%, 97.5\% for corresponding corpus sizes}
  \label{fig:training_dynamics}
\end{figure}

We use a pre-trained BERT model with 67M parameters as a backbone model (Table \ref{tab:models}). The hidden size is doubled because we concatenate subword embedding to each character embedding (Section \ref{neural_model}). 
Our  model was trained on 10M 
examples to get good spellchecking quality. See Fig. \ref{fig:training_dynamics} for training dynamics for different corpus sizes. The model was trained 
for 5 epochs on 8 GPUs with a global batch of $32*8$. We used the AdamW \cite{loshchilov2017decoupled} with an initial learning rate of 3e-5, and the CosineAnnealing learning scheduler with a warmup ratio of 0.1. 

We use two ASR models, Conformer-CTC and -RNNT (Table \ref{tab:models}) to provide initial  hypotheses. 
The testing procedure consists of the following steps:
\begin{enumerate}
\item Create a vocabulary for each ``user" (Section \ref{testing_datasets}).
\item Index each vocabulary using n-gram mappings (Section \ref{ngram_mappings_and_indexing}).
\item Transcribe all utterances with the baseline ASR model.
\item Loop through the transcriptions, split into fragments of 10-15 words in case of long sentences, and for each fragment retrieve the top 10 candidates from the corresponding index and perform inference with our neural model (Fig. \ref{fig:inference})
\item Apply post-processing and generate corrected or unchanged transcriptions.
\end{enumerate}

\subsection{Results}
Table \ref{tab:spellchecking_results} contains the results of the spellchecking model. 
We observe a substantial relative WER improvement (6.4-21.4\%) across three datasets from different domains over both baseline ASR models. Note that \textit{transducer} ASR model was not used in any stage of data preparation for the spellchecking model. Still, the post-correction gives a similar WER improvement, though recall and precision are slightly decreased (recall by 1.8-7.2\%, precision by 0.2-4.7\%), compared to the CTC model. We conclude that the SpellMapper generalizes well and can be applied without retraining on top of any ASR model for a given language. At the same time, retraining on the errors of specific ASR models is likely to improve performance further.
We observe a 20-33\% gap in recall between the candidate retrieval stage (top 10 recall) and the final recall. This means that our neural model and post-processing still lose a considerable amount of good candidates, so there is room for improvement in these stages.
The precision metrics between 77-87\% show that the model generates many false positives. This problem can be addressed in future research by sampling more training examples with false positives and adding trainable filters to the post-processing stage.
\footnote{
We cannot directly compare our results with \cite{Wang2022TowardsCS, sim2019personalization} as their evaluation was done on non-public datasets and no code is available.
}

\section{Conclusion}
This paper presents SpellMapper - a new approach for spellchecking customization of E2E ASR systems. We propose a new corpus-based algorithm for candidate retrieval, which does not rely on common letters, and provides 74-90\% recall with only ten candidates. We then reformulate the spellchecking problem as a simple task for BERT architecture without a decoder or custom neural blocks. 
We evaluate algorithm robustness and generality on three public datasets from different domains and with different ASR models.
In all cases, the model shows significant relative WER reduction (6.4-21.4\%) and allows to regain 49-68\% of maximum ”ideal” WER improvement with the given user vocabularies.
In future work, we would like to explore how to reduce false positives and increase the recall of correct candidates. We would also try to expand our approach to languages with highly inflectional morphology. 

\bibliographystyle{IEEEtran}
\bibliography{mybib}

\end{document}